%% file: root.tex
\documentclass[letterpaper, 10 pt, conference]{ieeeconf}  

\IEEEoverridecommandlockouts                              
\usepackage{graphics} 
\usepackage{epsfig} 
\usepackage{mathptmx} 
\usepackage{times} 
\usepackage{amsmath} 
\usepackage{amssymb}  
\usepackage{symbol}

\makeatletter
\let\NAT@parse\undefined
\makeatother
\usepackage[numbers,sort&compress]{natbib}

\title{\LARGE \bf
Jade: A Differentiable Physics Engine for Articulated Rigid Bodies with Intersection-Free Frictional Contact}

\author{Gang Yang$^{1}$, Siyuan Luo$^{2}$, and Lin Shao$^{1}$
\thanks{$^{1}$Gang Yang and Lin Shao are with the Department of Computer Science, National University of Singapore, Singapore.
        {\tt\small yg.matinal@gmail.com, and linshao@nus.edu.sg}}%
\thanks{$^{2}$Siyuan Luo is with the Department of Computer Science and Technology, Xi'an Jiaotong University, China.
        {\tt\small 312700@stu.xjtu.edu.cn}}%
}

\begin{document}

\maketitle
\thispagestyle{empty}
\pagestyle{empty}

\begin{abstract}
We present Jade, a differentiable physics engine for articulated rigid bodies. Jade models contacts as the Linear Complementarity Problem~(LCP). Compared to existing differentiable simulations, Jade offers features including intersection-free collision simulation and stable LCP solutions for multiple frictional contacts. We use continuous collision detection to detect the time of impact and adopt the backtracking strategy to prevent intersection between bodies with complex geometry shapes. We derive the gradient calculation to ensure the whole simulation process is differentiable under the backtracking mechanism. We modify the popular Dantzig algorithm to get valid solutions under multiple frictional contacts. We conduct extensive experiments to demonstrate the effectiveness of our differentiable physics simulation over a variety of contact-rich tasks. 
\end{abstract}

\section{Introduction}
\input{tex/intro}

\section{Related Work}
\input{tex/relatedwork}

\section{Preliminaries}
\input{tex/prelim.tex}

\section{Differentiable Simulation Design}
\input{tex/approach}

\section{Experiments}
\input{tex/exp}

\section{Conclusion}
\label{sec:conclusion}
\input{tex/con}

{\small
\bibliographystyle{IEEEtranN}
\bibliography{references}
}

\end{document}

%% file: tex/intro.tex
With recent advances in automatic differentiation methods~\cite{pytorch,team2016theano,hu2019taichi,bellCppAD,jax2018github,ceres-solver}, a number of differentiable physics engines, including rigid bodies~\cite{NEURIPS2018_842424a1,degrave2019differentiable}, soft bodies~\cite{hu2019taichi,hu2019chainqueen,jatavallabhula2021gradsim,geilinger2020add,du2021_diffpd}, cloth~\cite{NEURIPS2019_28f0b864,qiao2020scalable,li2022diffcloth,yu2023diffclothai}, articulated bodies~\cite{werling2021fast,58d18a9e94704e8bb63f5e2959de61d6,qiao2021efficient}, and fluids~\cite{um2020solver,wandel2020learning,holl2020learning,Takahashi_Liang_Qiao_Lin_2021}, have been developd for solving system identification and control problems.
These differentiable physics simulations provide differentiation operation to perform end-to-end optimization, which have been demonstrated to be effective for a broad range of application in robotics~\cite{degrave2019differentiable,hu2019taichi,NEURIPS2018_842424a1,geilinger2020add,qiao2020scalable,heiden2021neuralsim,escidoc:3221511,schenck2018spnets,NEURIPS2019_28f0b864,hu2019chainqueen,qiao2020scalable,holl2019learning,geilinger2020add,Zhu2023DiffLfD,lv2022sam,lv2022sagci}.

\begin{figure}[t!]
\centering
 \includegraphics[width=0.95\linewidth]{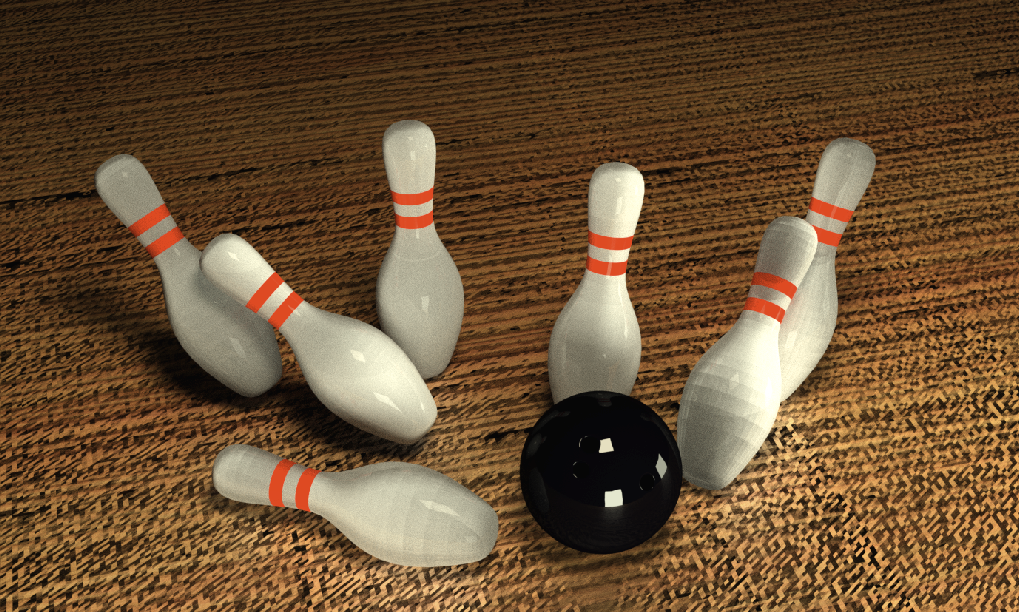}
 \caption{A bowling ball rolls on the floor and knocks pins off the ground at a speed of 10 meters per second. Our simulator under such this system is able to correctly calculate the collision moment and response force of every collision in the system to provide intersection-free results.}
\label{fig:teaser}
\end{figure}

Frictional contact is ubiquitous in robotics. Small violations of contact constraints introduce the penetration between articulated/rigid bodies, resulting in a significant deficiency in simulation accuracy and stability, especially for articulated rigid bodies. The majority of existing differentiable physics simulation perform poorly and results in penetration for contact-rich manipulation tasks requiring high-precision quality. \citet{chen2022midas} proposed a robotic simulation called \emph{Midas} for articulated rigid body under the IPC formulation~\cite{Li2020IPC} to provide penetration-free simulation. However, \emph{Midas} is not a differentiable simulation. 
~\citet{howell2022dojo} developed a differentiable physics simulation called Dojo to present a primal-dual interior-point to enforce no penetration. Dojo only supports primitive shapes and do not handle meshes for collision. 
Our differentiable simulation adopts a backtracking strategy to handle collision response which prevents the penetration and performs the gradient calculation. 

As a result, our differentiable simulation modeld contacts as the Linear Complementarity Problem~(LCP). Although different methods have been proposed for solving the LCP, including the popular Dantzig algorithm and Projected Gauss–Seidel~(PGS), we frequently observe the Dantzig or PGS fails to find a solution under the friction constraint, even in simple cases. (such as pushing a cube to slide along horizontal tables with frictions). In this work, we revised the Danzig algorithms to calculate the valid solution under the friction contacts. While tuning constraint forces to satisfy the complementary conditions, the original algorithm ignores the synchro variation of friction degree's upper/lower bound. So when the system has multiple correlated friction degrees and some are going to slide, the solution may go wrong. So we consider this issue and fix it. 

We provide an open-source implementation of our differentiable physics simulation, which we call Jade. Code and documentation will be released at our project website.

In summary, we make the following contributions:
\begin{itemize}
\item We adopt continuous collision detection to calculate the time of impact and use the backtracking strategy to prevent intersection between objects with complex shapes.
\item Under the backtracking mechanism, we derive symbolic differentiation rule for collision response and make the whole simulation differentiable.
\item We revise the LCP solver based on Dantzig's pivoting algorithm to handle strong-related friction situation.
\end{itemize}


%% file: tex/relatedwork.tex
In this section, we review related literature on key components in our approach, including contact modeling, numerical methods for LCP, and differentiable physics simulation.. We describe how we are different from previous work. 


\subsection{Contact Simulation}
Contact simulation for rigid bodies is one of the core components of physics simulations and is extensively studied in graphics and robotics.  Solving the contact impulses under frictions is inherently formulated as a nonlinear complementarity problem~(NCP). Dojo~\cite{howell2022dojo} developed
a primal-dual interior-point solver for the NCP problem. One common approach adopted by various simulations is to approximate the NCP as the Linear Complementarity Problem~(LCP) by approximating the friction cone with a polyhedral cone. A number of popular physics simulation engines integrated the LCP, such as ODE~\cite{ode:2008}, Bullet~\cite{coumans2021}, DART~\cite{lee2018dart}, Drake~\cite{drake} and PhysX~\cite{physX}. NCP/LCP formulations are sometimes referred to as hard contacts indicating the contact surfaces are rigid.

Unlike the hard contact formulations, Mujoco~\cite{conf/iros/TodorovET12} formulated the contact impulses calculation as a convex optimization problem by minimizing the post-collision kinetic energy. In mujoco, the complementarity condition can be violated, resulting in positive force and velocity values along the normal contact direction. Another line of contact simulations uses compliant models~\cite{giftthaler2017automatic,Carpentier2018AnalyticalDO,heiden2020neuralsim,geilinger2020add,freeman2021brax,warp2022}, assuming the contact surfaces can deform. Thus there are no impulses at the moment of collision, and there is no need to consider the linear complementarity problem. It is also easier to implement Coulomb friction at a compliant contact. However, these soft contact models allow the penetration to derive contact forces, which is not physically realistic. System parameters, such as object stiffness, might be difficult to tune for contact-rich manipulation tasks. Our differentiable physics simulation adopts the LCP formulation. To prevent the intersection, we adopt continuous collision detection and time of impact backtracking, which ensure the forward dynamics will stop and deal with collision response right before collision.

\subsection{Numerical Methods for Linear Complementarity Problems}
LCP is widely used as the model of contact impulses between rigid bodies in simulations. Multiple numerical algorithms have been proposed for solving the LCP. One line of approaches is to use the iterative algorithm, such as the Projected Gauss–Seidel~(PGS) type method. The LCP can be formulated as a minimization problem of a constrained convex QP problem, for which PGS is a suitable solver. 

Another type of popular solver is the pivoting algorithm~\cite{lemke}. Cottle and Dantzig~\cite{COTTLE1968103} presented a pivot algorithm for LCP called the Dantzig algorithm.  ~\citet{10.1145/192161.192168} extended the Dantzig algorithm to deal with friction and ~\citet{10.5555/2392643} provided a clear description. The pivoting methods can find an accurate solution to the LCP at a small computation cost for relatively small constraint sizes, whereas the iterative methods generally produce approximate solutions. Thus, it is practically reasonable to use the pivoting algorithm to limit computation time and switch to the iterative algorithm such as PGS~\cite{werling2021fast}. But in practice, we observe that the Dantzig's algorithm occasionally gives an invalid solution while solving frictional LCP. We find out the reason and present a modified solver based on the Dantzig's algorithm.

\subsection{Differentiable Physics Simulation}
Recently, great progress have been made in the field
of differentiable physics-based simulation with a number of differentiable physics engines for solving system identification and control problems. Differentiable simulations develop various gradient calculation approaches for learning, control, and inverse problems in physical systems. Here we review only differentiable simulations that support articulated rigid bodies.

~\citet{NEURIPS2018_842424a1} used the LCP formulation and derived gradients of a LCP solution with respect to input parameters based on implicit differentiation. A number of works~\cite{heiden2021neuralsim, degrave2019differentiable, qiao2021efficient} modeled contacts as the LCP with the PGS as the solver.  ~\citet{heiden2021neuralsim} and~\citet{degrave2019differentiable} leveraged existing automatic differentiation frameworks to get gradients whereas~\citet{qiao2021efficient} proposed a reverse version of the PGS solver using the adjoint method. Nimble~\cite{werling2021fast} computed analytical gradients through the LCP by exploiting the sparsity of the LCP solution. ~\citet{qiao2020scalable} leveraged the structure of contacts and grouped contacts into localized impact zones, where a QP is solved for each impact zone and the contact dynamics together with the conservation laws are not considered. ~\citet{geilinger2020add} proposed a differentiable physics engine with implicit forward integration and customized frictional contact model and developed a dynamics solver that is analytically differentiable. \citet{howell2022dojo} adopted the NCP formulation and develop
a primal-dual interior-point solver while obtaining the gradient based on the implicit-function theorem. Our differentiable physics simulation adopts the analytical gradients when time of impact backtracking is exported. We propose a sequential differentiation rule to calculate gradients when multiple collisions happen in single time inteval. 


%% file: tex/prelim.tex
This section contains background information about the continuous collision detection, which is a core component to prevent intersection between rigid bodies. 

\subsection{Continuous Collision Detection} 

In physical simulation, the motion of an object is modeled integrally through time discretization. In general, there are two ways to detect collisions, discrete collision detection (DCD) and continuous collision detection (CCD). Discrete collision detection computes and processes penetration based on each timestep, which can also be called remove the intersection state. In continuous collision detection for two meshes, edge-edge pairs and vertex-faces pair are recursively checked in each timestep to accurately calculate the time of collision and the magnitude of response force, so that CCD within a timestep is divided into multiple subtimesteps.

For vertex-face CCD, given a vertex $p$ and a face with three vertices $v_1, v_2, v_3$ at two distinct time steps $t^0$ and $ t^1$, finding if there exists a $t \in [t^0, t^1]$ that the vertex  $p$ is contained within the face supported by $v_1, v_2, v_3$. Similarly for edge-edge CCD the algorithm aims to check two moving edges $(p_1^t, p_2^t)$ and $(p_3^t, p_4^t)$ intersect. Now we introduce univariate CCD formulation.
A way of addressing the CCD problem is to convert it as a geometric observation: two primitives intersects if the four points, two pairs of edge's endpoints or a vertex and three triangle's vertices, are coplanar. So the problem becomes finding roots in a univariate cubic polynomial:
\begin{equation}
\begin{aligned}
f(t) = \left \langle n(t),q(t) \right \rangle = 0
\end{aligned}
\end{equation}

with $n(t) = (v_2(t) - v_1(t)) \times (v_3(t) - v_1(t))$ and $q(t) = p(t) - v_1(t)$ for vertex-face detection and $n(t) = (p_2(t) - p_1(t)) \times (p_4(t) - p_3(t)) $ and $q(t) = p_3(t) - p_1(t)$ for the edge-edge detection. Once the roots are identified, they need to be filtered, because not all roots correspond to actual collisions. For example, if two coplanar but disjoint edges are sliding at the same speed, they will continue to detect collisions that do not occur. Handling these cases, especially while accounting for floating point rounding, is very challenging. The root finding formula is the method we often use to solve cubic equations of one variable. However, due to its accuracy error, Super-sampling or Bisection method\cite{bisection} are developed to handle it, which could strike a balance between efficiency and accuracy.

\subsection{Linear Complementarity Problem}
Linear complementary problem (LCP) is a classic model for rigid-body contact constraints with problem parameters $\A$ and $\be$, where $\A$ is symmetric and positive semidefinite and reflects the masses and contact geometries of the bodies and $\be$ is a vector in the column space of $\A$ and reflects the external and inertial forces in the system. From Newton's law, we have the dynamic function: $\acc = \A\f + \be$, where $\f$ is the vector of normal forces on each contacts, and $\acc$ is the relative acceleration between two objects with contact. This equation has infinite solutions because $\acc$ is unknown. However, from contact constraints and conservation laws, the contact force $f_i$ and relative acceleration $a_i$ of contact point $i$ should satisfy some conditions, so that we could iteratively solve $\f$. As for frictionless contact $i$, we have:

\begin{equation}
\label{eqn:lcp}
a_i \geq 0 , f_i \geq 0 \text{ and } f_i a_i = 0
\end{equation}

The solution $\f = \text{LCP}(\A, \be)$ can be solved iteratively by the Dantzig’s algorithm \cite{10.1145/192161.192168}.

%% file: tex/approach.tex
 Simulators use positions and velocities at each discrete timestep to calculate the forward dynamics and the backward gradients. However, only considering the states at fixed time stamp might result in the intersection between rigid bodies, especially when the object's velocity in simulation is too fast or the timestep is set too large. To solve this problem, this work exports the continous collision detection~(CCD) module and the time-of-impact (TOI) backtracking strategy into forward dynamics. We then design the corresponding differentiation rule for new dynamics. Besides the designing intersection-free differentiable simulation, we occasionally notice invalid contact solutions from current LCP solver and provide a modified solver to prevent invalid solutions.
 

\subsection{Differentiable Simulator with DCD}

\subsubsection{Forward Dynamics}
An articulated rigid body with discrete timestep can be thought of as a simple function $[\q_{t+1}, \dq_{t+1}] = P(\q_t, \dq_t, \m_t, \tf_t, \dt)$, where $P(\cdot)$ is the collision-free forward dynamics function. The collision detections and responses are all dealt at the end of discrete timesteps, so that $P(\cdot)$ is continous and linear on $\dt$.
The inputs of $P(\cdot)$ are current position $\q_t$, velocity $\dq_t$, control forces $\tf_t$, inertial properties $\m_t$ and timestep $\dt$ as input, and returns the next state, $\q_{t+1}$ and $\dq_{t+1}$. In this work, we use explicit Euler integral method to formulate forward dynamics:  

\begin{equation}
\label{eqn:dynamics}
P(\cdot): 
\left \{
\begin{aligned}
    \dq_{t+1} &= \dq_t + \Minv_t \z_t \\
    \q_{t+1} &= \q_t + \dt \dq_t \\
    \z_t &\equiv \dt(\tf_t - \corio_t) + \J_t^{T}\f_t \\
    \f_t &= \text{LCP}(\A_t, \be_t) \\
    \A_t &= \J_t\Minv_t\Jt_t \\
    \be_t &= \J_t(\dq_t + \dt\Minv_t(\tf_t - \corio_t)) \\
\end{aligned}
\right.
\end{equation}

where $\M$ is the mass matrix, $\Delta t$ is the timestep, $\corio$ is Coriolis and gravitational force, $\f$ is the contact force, $\J$ is the contact Jacobian matrix and $\z$ is the total impulse. In our notation, $\q$, $\dq$ and $\tf$ are all expressed in generalized coordinates, describing the relative motion of joints in articulated rigid body system. Thus joint constraint conditions are automatically satisfied, and we mainly need to care about contact constraints, which is represented by a linear complementarity problem (LCP) that we will introduce later. We calculate the contact force $\f$ by solving LCP.

\subsubsection{Backward Differentiation}
Now that we have the full analytical formula of $P(\q_t, \dq_t, \m_t, \tf_t, \dt)$, it's able to directly differentiate it and calculate full gradients:

\begin{equation}
\label{eqn:dP1}
\left\{
\begin{aligned}
    \frac{\partial \q_{t+1}}{\partial \q_t} =& \I \\
    \frac{\partial \q_{t+1}}{\partial \dq_t} =& \dt\I \\
    \frac{\partial \q_{t+1}}{\partial \dt} =& \dq_t
\end{aligned} 
\right.
\end{equation}
    
\begin{equation}
\label{eqn:dP2}
\left\{
\begin{aligned}
    \frac{\partial \dq_{t+1}}{\partial \q_t} =& \frac{\partial \M_t^{-1}\z_t}{\partial \q_t} + \M_t^{-1}(-\dt \frac{\partial \corio_t}{\partial \q_t} + \frac{\partial \J_t^T}{\partial \q_t}\f_t) \\
    &+ \J_t^T\frac{\partial \f_t}{\partial \q_t}  \\
    \frac{\partial \dq_{t+1}}{\partial \dq_t} =& \I + \M_t^{-1}(-\dt \frac{\partial \corio_t}{\partial \dq_t} + \J_t^T\frac{\partial \f_t}{\partial \dq_t}) \\
    \frac{\partial \dq_{t+1}}{\partial \tf_t} =& \M_t^{-1}(\dt \I + \J_t^T \frac{\partial \f_t}{\partial \tf_t}) \\
    \frac{\partial \dq_{t+1}}{\partial \m} =& \frac{\partial \M_t^{-1} \z_t}{\partial \m} + \M_t^{-1}(-\dt \frac{\partial \corio_t}{\partial \m} + \J_t^T \frac{\partial \f_t}{\partial \m}) \\
    \frac{\partial \dq_{t+1}}{\partial \dt} =& \M_t^{-1}(\tf_t - \corio_t) + \J_t^{T}\frac{\partial \f_t}{\partial \dt} 
\end{aligned}
\right.
\end{equation}

where $\I$ is the identity matrix. However, DCD method could cause severe intersection, especially when the objects move fast. To prevent intersection, our differentiable simulator is designed to adopt continuous collision detection (CCD) instead.

\subsection{Differentiable Simulator with CCD}
\begin{figure}[htb!]
  \begin{center}
   \includegraphics[width=1.0 \linewidth]{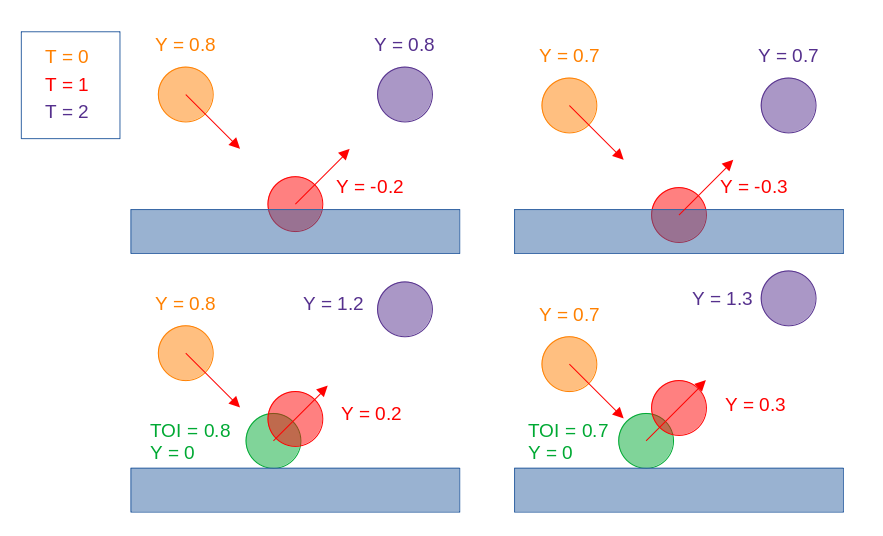}
  \end{center}
      \caption{Elastic collision between ball and floor. Top row is the DCD case and bottom row is the CCD case. Orange, red and purple balls mean the positions of T = 0, 1, 2 and green ball means the position at time of impact. Each case, compare the left figure and right figure, we can calculate the gradient $\frac{\partial Y_2}{\partial Y_0}$. DCD gets the wrong result 1 while CCD gets the correct result -1.}
\label{fig:toi}
\end{figure}

\subsubsection{Time-of-impact Backtracking}

In simulation design, we use the time-of-impact (TOI) to record the exact time when two object collide. Continuous collision detection~(CCD) is used to predict TOI. If TOI is less than the timestep $\dt$, we know there will be collision within this time interval. Then we should track the motion back to TOI (the moment collision happens) and deal with collision response, and then run over the remained time. One TOI backtracking divides an entire timestep into two parts and solves each motion with different velocity because the collision changes the velocity. Fig.\ref{fig:toi} shows an example that a ball collides with floor elastically. The top row simply solves collision at discrete time, while bottom row adopts TOI backtracking. From this figure, we see that the motion with TOI backtracking is more accurate and prevents intersection.
Adopting CCD into differentiable simulator not only gives better forward dynamics but also corrects wrong gradients. We then construct the dynamics equations with TOI backtracking for general complex environment with multiple collisions.

As mentioned in previous section, the forward dynamics without collision can be described as $[\q_{t+1}, \dq_{t+1}] = P(\q_t, \dq_t, \m_t, \tf_t)$, if timestep $dt$ is a constant. But when a collision is detected by CCD at TOI, the whole forward dynamics would be divided into three parts: two collision-free propagation before and after collision, and one collision response at TOI. Thus timestep $dt$ should be taken as a variable and collision-free propagation becomes its function $[\q_{t+1}, \dq_{t+1}] = P(\q_t, \dq_t, \m_t, \tf_t, \dt)$. We use $[\q^-, \dq^-]$ and $[\q^+, \dq^+]$ to denote the state right before and after collision, and use $\dt_c$ to denote TOI. So the full forward dynamics $F(\cdot)$ over $dt$ becomes:

\begin{equation}
F(\cdot): 
\left \{
\begin{aligned}
\relax [\q^-, \dq^-] &= P(\q_t, \dq_t, \m_t, \tf_t, \dt_c) \\
[\q^+, \dq^+] &= C(\q^-, \dq^-, \m_t) \\
[\q_{t+1}, \dq_{t+1}] &= P(\q^+, \dq^+, \m_t, \tf_t, \dt - \dt_c) \\
\dt_c &= T(\q_t, \dq_t)
\end{aligned}
\right .
\end{equation}

where $C(\q, \dq, \m)$ is the collision response function and $T(\q, \dq)$ is the collision detection function. After deriving $C(\cdot)$, $T(\cdot)$ and their differentiations, we can combine with $P(\cdot)$ and get the full formula of forward and backward dynamics. 

\subsubsection{Collision Response}

Although collision usually happens between one pair of objects, all objects linking to them through chains of constraints (joint and contact) should be considered when we calculate the response impulse. In physics, the impulse conductive velocity of rigid body is infinity, so the collision response of all relevant objects are solved simultaneously. Making use of the dynamic properties of articulated rigid body, we will show the collision response function $C(\q, \dq, \m)$ is equivalent to the collision-free propagation function $P(\cdot)$ under a variable substitution \cite{collision_lcp}. Then we can directly adapt previous results derived from $P(\cdot)$. 

To calculate collision impulses, we formulate the problem as a LCP. We use $\vel^-$ and $\vel^+$ to denote the relative velocities of all contact points before and after collision, where $\vel > 0$ means separating, $\vel < 0$ means approaching and $\vel = 0$ means relatively static. 

\begin{figure}[ht]
  \begin{center}
   \includegraphics[width=1\linewidth]{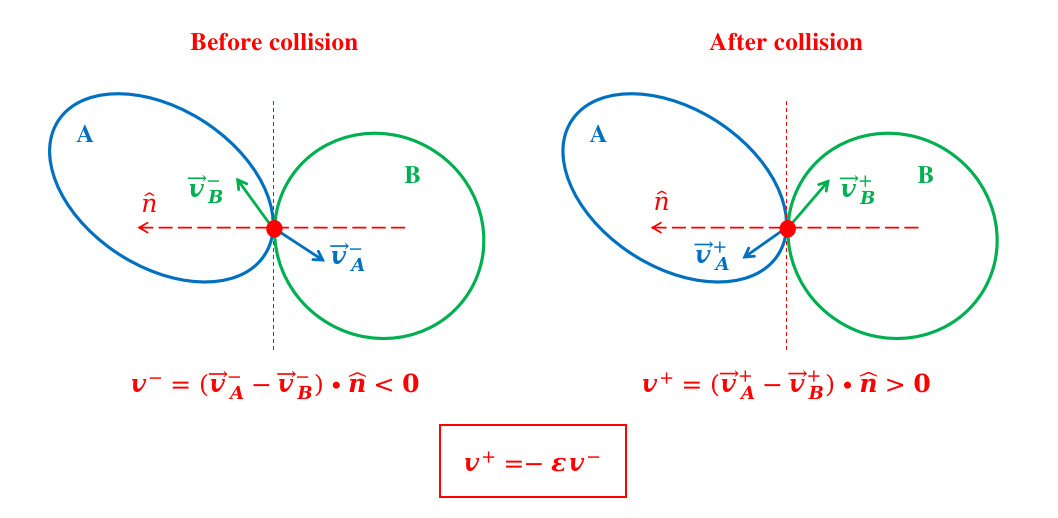}
  \end{center}
      \caption{The relative velocity change after collision.}
\label{fig:collision}
\end{figure}

\begin{itemize}
    \item For collision points, the relative velocities obey the elastic collision rule as shown in Fig.\ref{fig:collision}: $\vel^+ = -\epsilon \vel^-$, where $\epsilon \geq 0$ is the restitution coefficient. Since the relative velocity difference $\vel^+ - \vel^- = - (1+\epsilon) \vel^- > 0$, the collision impulse should also be positive, $\f > 0$.
    \item For non-collision contact points, same as the common LCP, we have $v^- = 0$, $v^+ \geq 0$, $\f \geq 0$ and $\f \vel^+ = 0$.
\end{itemize}

We can combine the two cases and derive a complementary condition just like what we know in LCP:

\begin{equation}
v^+ + \epsilon v^- \geq 0 , f_i \geq 0 \text{ and } f_i (v^+_i + \epsilon_i v^-_i) = 0
\end{equation}

\begin{figure}[ht]
  \begin{center}
   \includegraphics[width=1\linewidth]{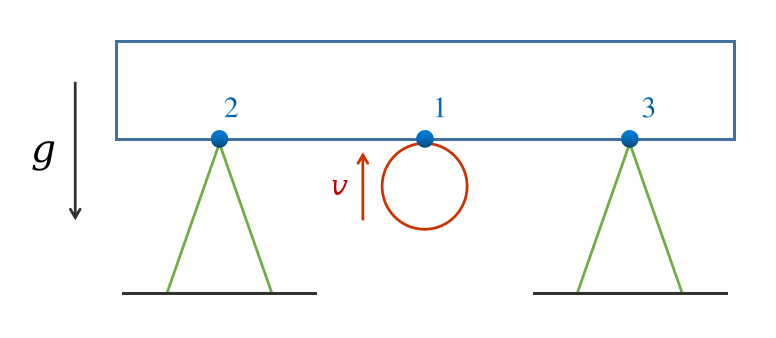}
  \end{center}
      \caption{The relative velocity change at all contact points.}
\label{fig:collision_ineq}
\end{figure}

For example, in Fig.\ref{fig:collision_ineq}, the blue rectangle rests on two green triangles in gravity at the beginning. One red circle moves upwards and collides with the blue rectangle. So we have $v^-_1 = -\|\vec{v}\|, v^-_2 = v^-_3 = 0$ for the 3 contact points before collision, and $v^+_1 = \epsilon \|\vec{v}\|, v^-_2 \geq 0, v^-_3 \geq 0$ after collision, which is just $v^+_i + \epsilon v^-_i \geq 0$.

The Newton's law in contact space again gives $\A\f = \vel^+ - \vel^-$, which can be rewritten as $\A \f + \be = \vel^+ + \epsilon \vel^-$, where $\be = (1 + \epsilon) \vel^-$. So the collision response problem is still a LCP. Since collision happens in an instant, all the impulses from external force and gravity equals to zero. We can regard the collision response function as a variation of propagation function by setting $\dt = 0$, i.e. $C(\q, \dq, \m) = P(\q, \dq, \m, \tf = 0, \dt = 0)$ with a modified LCP parameter $\be = (1 + \epsilon) \J\dq$:

\begin{equation}
C(\cdot): 
\left \{
\begin{aligned}
    \q^+ &= \q^- \\
    \dq^+ &= \dq^- + \Minv \Jt\f \\
    \f &= \text{LCP}(\A_c, \be_c)\\
    \A_c &= \J\Minv\Jt \\
    \be_c &= (1 + \epsilon) \J\dq^-
\end{aligned}
\right.
\end{equation}

The gradients of $C(\cdot)$ are directly obtained from Eq.\ref{eqn:dP1} and Eq.\ref{eqn:dP2} by substituing $C(\q, \dq, \m) = P(\q, \dq, \m, \tf = 0, \dt = 0)$ and $\f = \text{LCP}(\A_c, \be_c)$

\subsubsection{Differentiate TOI}

\begin{figure}[ht]
  \begin{center}
   \includegraphics[width=0.8\linewidth]{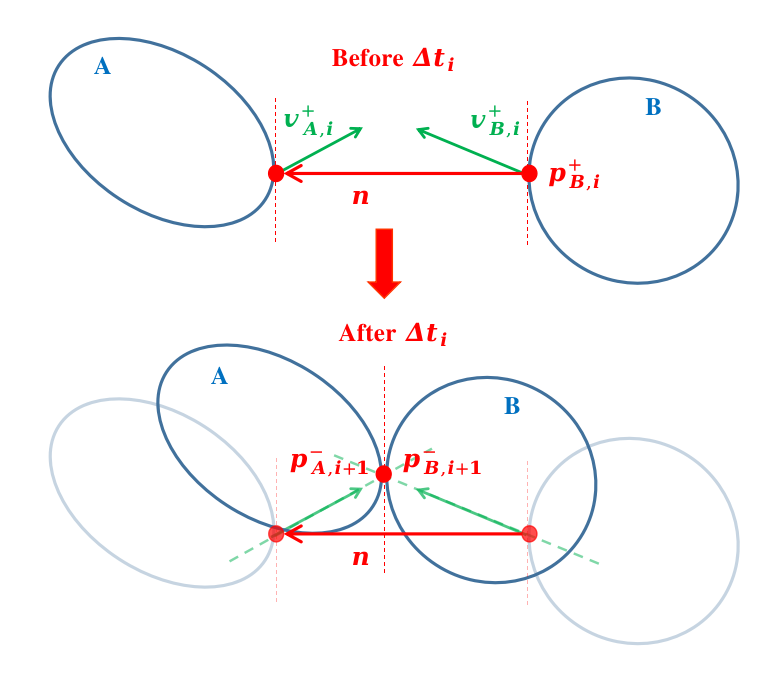}
  \end{center}
      \caption{The motion before two objects collide.}
\label{fig:delta_t}
\end{figure}

The time of impact is obtained from an iterative calculus in CCD, which is hard to directly differentiate step by step. So we another expression of TOI assuming we have already known the collision point (which is exactly solved from CCD) and calculate gradients. To give a clear picture of what's going on when we differentiate TOI, we show a colliding pair's position relation at two timestamps in Fig.\ref{fig:delta_t}. Then TOI is just the time for relative distance $|\n|$ decreasing to zero, $T(\cdot)$ and its gradients are:

\begin{equation}
T: \dt_c = \frac{\|\n\|}{(\vel^B_t-\vel^A_t) \cdot \n} = \frac{\|\n\|}{\J^-_n\dq_t}
\end{equation}

\begin{equation}
\partial T: 
\left \{
\begin{aligned}
    \frac{\partial \dt_c}{\partial \q_t} &= \frac{\J^-_n}{\J^-_n\dq_t} \\
    \frac{\partial \dt_c}{\partial \dq_t} &= \frac{\J^-_n}{\J^-_n\dq_t} \dt
\end{aligned}
\right.
\end{equation}

where $\J^-_n$ is the Jacobian of collision point's normal direction component at TOI.

\subsection{Dantzig's Algorithm for LCP}

Unlike formulating LCP as a quadratic program (QP) and solving it in an optimisation manner, Dantzig's algorithm \cite{10.1145/192161.192168} reconstructs the nonlinear complementary conditions in Eq.\ref{eqn:lcp} into a set of contact classes $\{C_1, C_2, ... , C_s\}$ where each class refers to one linear condition. The origin LCP is then divided into a set of linear equations $\{E_1, E_2, ... , E_s\}$ which are easy to solve.

\subsubsection{Dantzig’s Algorithm without Friction}

\begin{figure}[htb!]
  \begin{center}
   \includegraphics[width=0.6\linewidth]{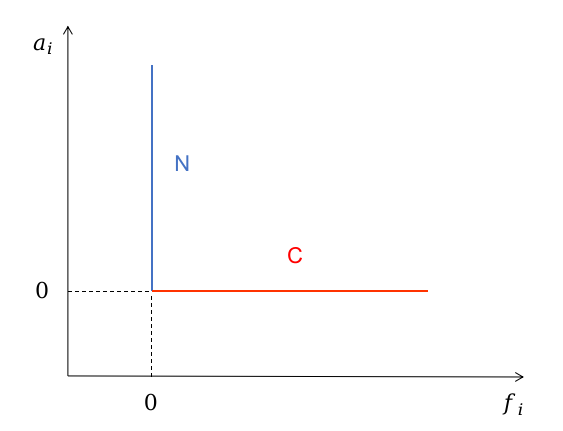}
  \end{center}
      \caption{LCP solution zone without friction} 
\label{fig:frictionless_zone}
\end{figure}

If there's no friction, the complementary condition for contact $i$ is Eq.\ref{eqn:lcp}, which can be visualized by regarding $(f_i, a_i)$ as a 2D point. The valid solutions $(f_i, a_i)$ satisfying the complementary condition should lie on a right-angle, as shown in Fig.\ref{fig:frictionless_zone}. Since the valid solutions has two linear segements, the contact classes are $\{C_1 = C, C_2 = N\}$, where class $C$ means clamping contact with zero acceleration and class $N$ means non-clamping or separating contact with zero force.

After rearranging the indice according to classes, $\f = [\f_C, \f_N]$, we only need to solve two linear equations $\{E_C, E_N\}$:

\begin{equation}
\begin{bmatrix}
    \acc_C \\
    \acc_N
\end{bmatrix}
=
\begin{bmatrix}
    \A_{CC} & \A_{CN} \\
    \A_{NC} & \A_{NN} 
\end{bmatrix}
\begin{bmatrix}
    \f_C \\
    \f_N
\end{bmatrix}
+
\begin{bmatrix}
    \be_C \\
    \be_N
\end{bmatrix}
\end{equation}

\begin{equation}
\left\{
\begin{aligned}
    E_C: &\A_{CC} \f_C + \be_C = 0 \\
    E_N: &\f_N = 0
\end{aligned}
\right.
\end{equation}

\subsubsection{Dantzig’s Algorithm with Friction}

When contact points have friction, the previous complementary condition will change because of the Coulomb friction law. If $\f_i$ is a friction force component, let $f_{n_i}$ denotes its normal force component and $\mu_i$ denotes the friction coefficient, the complementary condition and classification are:

\begin{equation}
\begin{aligned}
|\f_i| \leq \mu \f_{n_i}, \acc_i \f_i \leq 0 \\
\acc_i(\mu \f_{n_i} - |\f_i|) = 0
\end{aligned}
\end{equation}

\begin{equation}
\left\{
\begin{aligned}
F: &\acc_i = 0 , - \mu \f_{n_i} < \f_i < \mu_i \f_{n_i} \\
H: &\acc_i < 0, \f_i = \mu \f_{n_i} \\
L: &\acc_i > 0, \f_i = - \mu \f_{n_i}
\end{aligned}
\right.
\end{equation}

\begin{figure}[htb!]
  \begin{center}
   \includegraphics[width=0.6\linewidth]{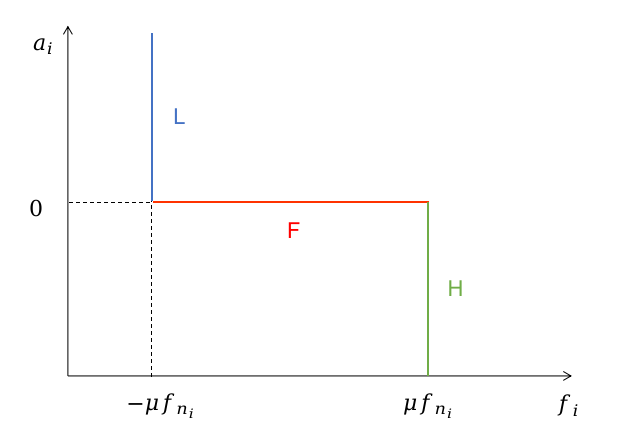}
  \end{center}
      \caption{LCP solution zone with static friction.} 
\label{fig:friction_zone}
\end{figure}

Similarly, We can visualize the valid solutions of frictional $(f_i, a_i)$ in Fig.\ref{fig:friction_zone}. For convenience, we divide class $C$ into $\{C_F, C_H, C_L\}$ indicating the class of its friction component, and the linear equations are:

\begin{equation}
\left\{
\begin{aligned}
    E_C: &(\A_{C C_H} + \mu \A_{C H}) \f_{C_H} + (\A_{C C_H} - \mu \A_{C L}) \f_{C_L} \\
        &+ \A_{C C_F} \f_{C_F} + \A_{C F} \f_F + \be_C = 0 \\
    E_N: &\f_N = 0 \\
    E_F: &(\A_{F C_H} + \mu \A_{F H}) \f_{C_H} + (\A_{F C_H} - \mu \A_{F L}) \f_{C_L} \\
        &+ \A_{F C_F} \f_{C_F} + \A_{F F} \f_F + \be_F = 0 \\
    E_H: &\f_H = \mu \f_{C_H} \\
    E_L: &\f_L = - \mu \f_{C_L}
\end{aligned}
\right.
\end{equation}

\subsubsection{Iterative Classification}

From previous sections, we know the key point of solving LCP is the contact classification. As long as we have correctly classified all force components, solving linear equations is simple. In original Dantzig's algorithm, the contact classification is implemented iteratively as shown in Alg.\ref{alg:Dantzig}.

\begin{algorithm}
\caption{Dantzig’s Algorithm}\label{alg:Dantzig}
\begin{algorithmic}[1] 

\State $D$ denotes the dimension of contact forces
\State $S_i$ denotes the solution set of $(\f_i, \acc_i)$ for $i \in [1, D]$
\State $C_1, C_2, ... , C_s \gets \emptyset$
\State $\f \gets \0$

\State $k \gets 0$
\While{$k < D$ }
    \State $\acc_{k+1} \gets \A_{k+1,1:k} \f_{1:k} + \be_{k+1}$
    \While{$(\f_{k+1}, \acc_{k+1}) \notin S_{k+1}$}
        \State Increase $|\f_{k+1}|$
        \State Update $(\f_{1:k}, \acc_{1:k})$ keeping classes unchanged
        \State Some $(\f_i, \acc_i)$ is at the boundary between \newline
        \hspace*{5em}current class $C_i$ and its neighbor $C^{\prime}_i$
        \State Assign $i$ from $C_i$ to $C^{\prime}_i$
    \EndWhile
    \State Assign $k+1$ to corresponding $C_x \in \{C_1, C_2, ... , C_s\}$
    \State $k \gets k+1$
\EndWhile
\State Solve linear equations $\{E_1, E_2, ... , E_s\}$

\end{algorithmic}
\end{algorithm}

\subsubsection{Failure Issues and Our Improvement}

In practice, however, the Dantzig's LCP solver implemented by ODE sometimes fails for two reasons:

\begin{itemize}
    \item \textbf{Implement Issue}: Ignore the correlation between friction limit and normal force while updating forces. For example, when a frictional force $\f_i$ in class $F$ is going to reach the boundary between $F$ and $H$, current solver calculates the max step as $s_i = \frac{\mu \f_{n_i} - \f_i}{\Delta \f_i}$. However, $\f_{n_i}$ is not a constant value as $\f$ varies and $\f_i$'s upper bound $\mu f_{n_i}$ is not constant either. The true step should be $s_i = \frac{\mu \f_{n_i} - \f_i}{\Delta \f_i - \mu \Delta \f_{n_i}}$ and we fix this issue in our new solver.
    \item \textbf{Theoretical Issue}: No insurance of the convergence while increasing new added $|\f_{k+1}|$. In frictional case, mathematically there is no proof that continuously increasing $\f_{k+1}$ in one direction (positive or negative) will lead $1:k+1$ assigned to correct classes. Infact, sometimes this might cause infite loop and algorithm does not converge. So our new solver take the original line search as an initial try, and apply ergodic search from the initial try when loop is detected until the correct class assignment is found.
\end{itemize}

%% file: tex/exp.tex
Our experiments focus on evaluating the following questions: 1) Can our differentiable physics simulation produce penetration-free simulation results for contact-rich manipulation tasks? 2) How accurate is our differential simulation gradient calculation? 3) How stable is our proposed LCP solver?

\subsection{Precise Collision Detection and Response}

\begin{figure}[htb!]
  \begin{center}
   \includegraphics[width=1.0\linewidth]{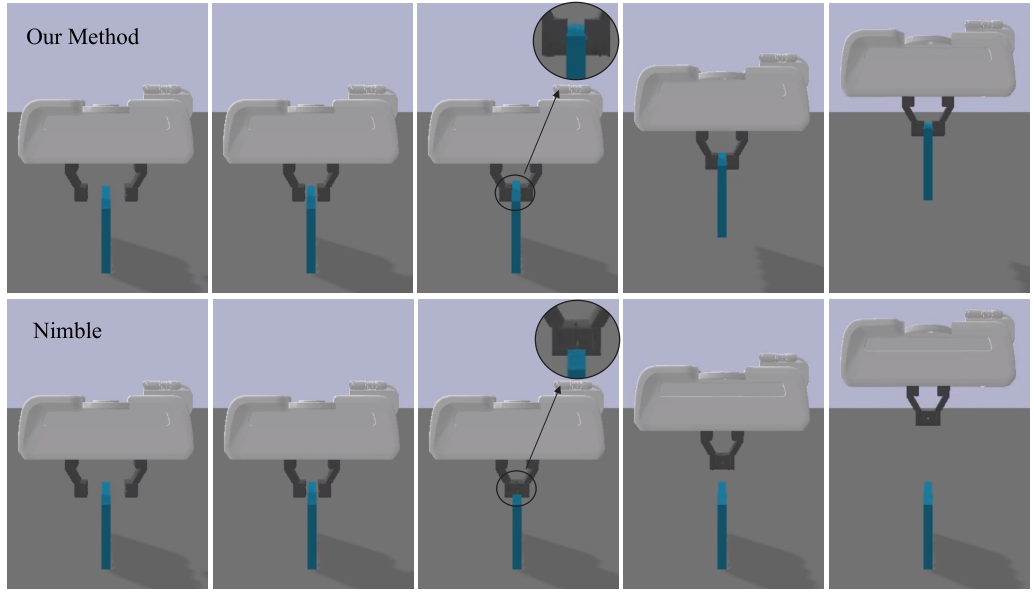}
  \end{center}
      \caption{We use grippers to pick up a thin plate, the thickness of the plate is 0.02m, we compare our simulator with Nimble. Grippers in Nimble penetrate the plate, and our simulator can tightly hold on the plate and pick it up.}
\label{fig:exp1_big}
\end{figure}

\begin{figure}[htb!]
  \begin{center}
   \includegraphics[width=1.0\linewidth]{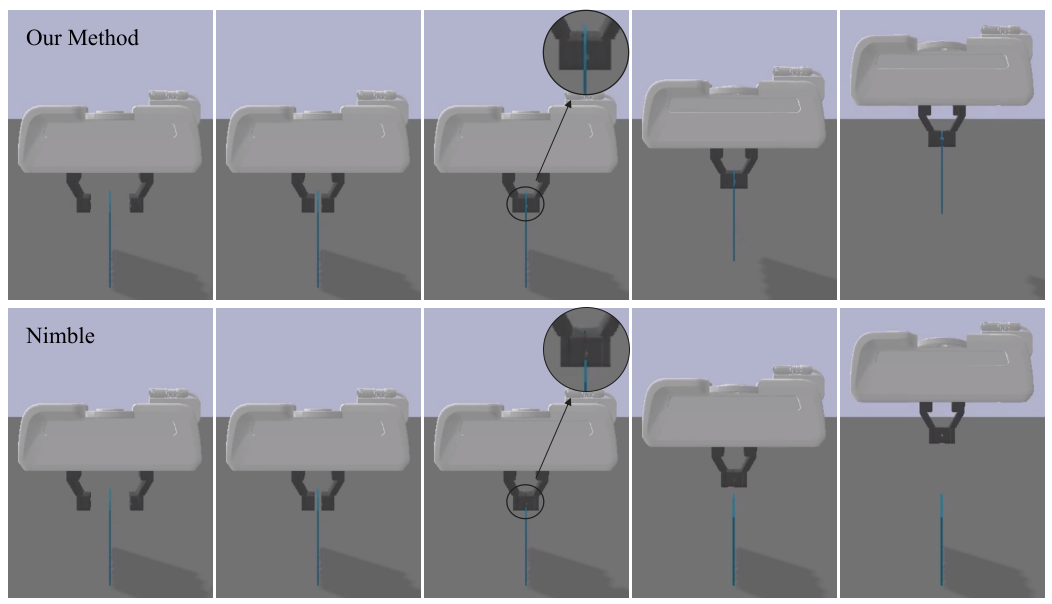}
  \end{center}
      \caption{We change the thickness of the plate into 0.002m, and do the same task with Fig.\ref{fig:exp1_big}. Our simulator can still solve this problem, and we compare our result with Nimble.}
\label{fig:exp1_small}
\end{figure}

Picking up thin-shell objects is challenging for robotic simulations because the intersection easily happens between grippers and objects such as plates and boards. In this experiment, we control a Franka robot's gripper with force to pick up a thin plate. The thickness of this plate is about 0.02m. The task requires the parallel-jaw gripper to close the fingers to grasp the object and then control the gripper to move upward. For details implementation, we set the time step as 0.01 second and the friction coefficient as 0.8. Our simulation lasted for 100 time steps. For the first 20 timesteps, we keep the base still and give the gripper an inward clamping force, and the gripper keeps moving until it grips the sheet. For the next 80 timesteps, we continuously give the gripper inward force and simultaneously move the base upward. The magnitude of the clamping force is always 1N, the mass of the thin plate is 0.1kg, and the gravitational acceleration of the system is $8m/s^{2}$.

We compare our result with other robotics simulators, such as Nimble~\cite{werling2021fast}. Because of our high-precision CCD, accurate collision response, and contact calculation of LCP, our gripper can hold and move the plate tightly. Even if we apply a larger force or bigger initial velocity on the gripper, our simulator can still solve the collision. For example, we increase the initial force of the gripper to 10N, so that when the gripper and the thin plate collide, their relative speed is larger. There is still no penetration in our simulation, which shows the effectiveness of our simulation. Then we move up the base of the robot, when the sheet left the ground, we reduced the clamping force back to 1N, so the friction force is just the same as gravity by correcting the clamping force of the gripper, and the thin plate still does not fall, which shows the accuracy of our friction calculation. However, when simulating the same task with Nimble\cite{werling2021fast}, the larger gripper velocity will cause the tunneling problem\cite{tunneling}. When the gripper opens, the plate is unable to drop on the floor. We also change the thickness of plane from 0.02m to 0.002m, our simulator still occur no penetration. Fig.\ref{fig:exp1_big} illustrates the experiment results for 0.02m thickness sheet, and Fig.\ref{fig:exp1_small} for 0.002m thickness sheet. We put videos recording this experiment on our project website.

This experiment demonstrates the accuracy of collision detection and response in our forward robotics simulation under hard conditions, as well as the calculation of friction force.

\subsection{Gradient Calculation for grid-based object}
In this experiment, we design the task of learning optimal control with a two-ball collision in a plane.
We compute our gradient calculation with other differentiable simulators including Nimble\cite{werling2021fast}, Brax~\cite{freeman2021brax} and Diff-taichi~\cite{hu2019difftaichi}.

\begin{figure}[htb!]
  \begin{center}
   \includegraphics[width=0.9\linewidth]{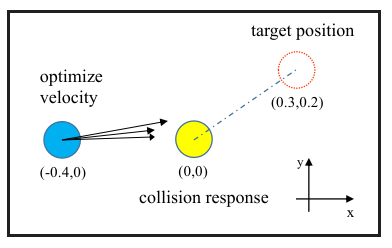}
  \end{center}
      \caption{We place two balls on a table, a red one and a blue one.The friction coefficient of the system is zero. In this task, we set the initial velocity of the blue ball, and hit the yellow ball to the position of red dashed line. We use our simulator to optimize the initial velocity of the blue ball.}
\label{fig:exp2_demo}
\end{figure}

We design a task to verify gradient calculation by setting the same task and comparing the optimize results. A blue ball and a yellow ball are placed on a table which has no friction, and the radius of the two small balls is 0.05m in Fig.\ref{fig:exp2_demo}. We take the center of mass of the yellow ball as the origin, and the table top is the plane to establish a planar rectangular coordinate system. The blue ball has a centroid coordinate of [-0.4,0], and the yellow ball has a centroid coordinate of [0,0]. By optimizing the initial velocity of the blue ball, the blue ball hits the yellow ball such that the yellow ball moves to the specified position $p_{target} $ = [0.3,0.2] in the simulation process. The whole simulation process lasts 0.2s, with a total of 20 timesteps. For each differentiable simulator, we optimize for 1000 epochs. The initial velocity of the blue ball in the first epoch is (4,-0.1). We mark the target position for yellow ball as $p_{final}$, the loss function for all simulator is designed as $ Loss = \left \| p_{final} - p_{target}  \right \|_2^2  * 10$.

\begin{figure}[htb!]
  \begin{center}
   \includegraphics[width=0.87\linewidth]{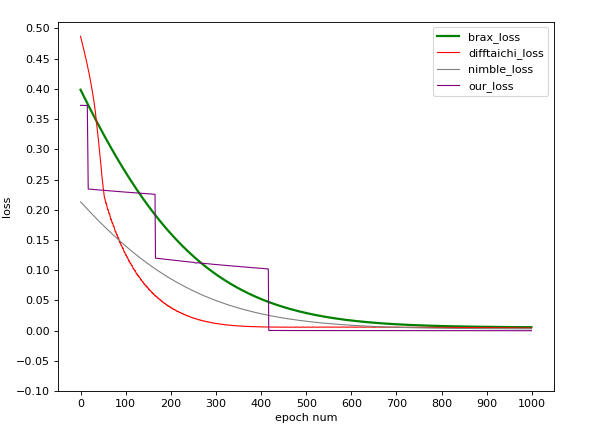}
  \end{center}
      \caption{The X-axis represents the number of optimization epochs and the Y-axis represents the loss. Among them, Brax has the slowest decline and our method has the lowest loss.}
\label{fig:exp2_draw}
\end{figure}

We also compare the loss and optimize result with other simulators which support the gradient calculatio in Fig.~\ref{fig:exp2_draw}.
For Nimble Simulator, we set all the parameters same as our simulation. For Brax and Diff-taichi, what we're actually optimizing is the inital forces, which acts on the first timestep integrate to approximately optimize the initial state velocity.The loss function and physical parameters are kept consistent in all simulators.

\begin{table}[tbh!]
\caption{Quantitative results}
\begin{center}
\begin{tabular}{c|ccc}
\hline\hline
\bf{engine} &  \bf{Final Position} & \bf{Position Error} & \bf Initial Velocity\\
\hline
Brax(PBD) & (0.31035, 0.1398) & 0.05043 & (4, -0.5621)  \\ 
Diff-Taichi & (0.3172, 0.1642) & 0.01577 & (4, -0.5715)\\ 
Nimble & (0.3008, 0.1751) & 0.02491 & (4, -0.5819)\\ 
Ours & (0.3012, 0.2010)& 0.00156 & (4, -0.6285)\\ 
Analytical & (0.3, 0.2) & 0 & (4,-0.7003)\\
\hline
\hline
\end{tabular}
\end{center}
\label{tab:BP}
\end{table}

We report the results in Tab.~\ref{tab:BP}. In Nimble, Brax and Diff-taichi simulator, the balls are initialized with primitives. However, in our simulator, we initialize the balls are meshes. Although our meshes of the ball itself has the largest  error in terms of accuracy, but our results are among the best. We compare the error in two dimensions, one is the descent trajectory of the loss function, and the other is the distance between the optimized yellow ball position and the target position. In terms of loss function, Brax's loss decreases slowly, Diff-taichi's and Nimble's loss go down faster. Our loss shows an approximate step-like descent, which is due to the rounding approximation of gradient calculation caused by coarse detection in CCD detection, and has no effect on the optimization results. In terms of the distance between optimized yellow ball position and the target position, as shown in Tab.~\ref{tab:BP}, the Error was calculated by the Euclidean distance between Final Position and Target Position. Our error is more than twenty times smaller than Nimble. The experiment shown above indicates that the gradients provided by \textbf{Jade} are more accurate than other differentiable primitive-based simulators, and compared with the analytical result our optimized results are more reliable. We put videos recording this experiment on our project website. 


\subsection{Invalid LCP Solution}

\begin{figure}[htb!]
 \begin{center}
  \includegraphics[width=0.87\linewidth]{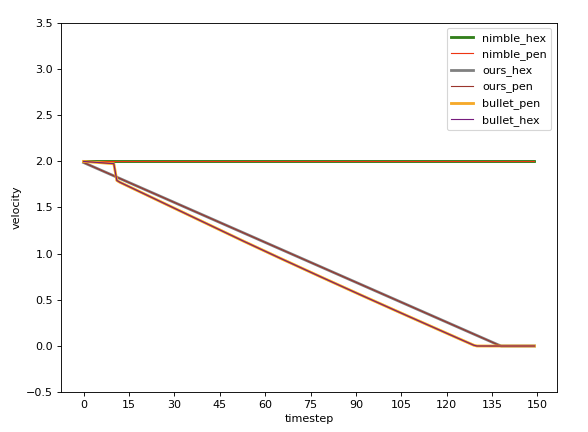}
 \end{center}
     \caption{The X-axis represents the number of timesteps, each time step is 0.01 second, and the Y-axis represents the velocity for prisms. With friction, Nimble's results show the friction dropped.}
\label{fig:slide_table}
\end{figure}

\begin{figure*}
    \centering
    \subfigure[Pushing cube in Bullet]
    {
        \includegraphics[height=1.8in]{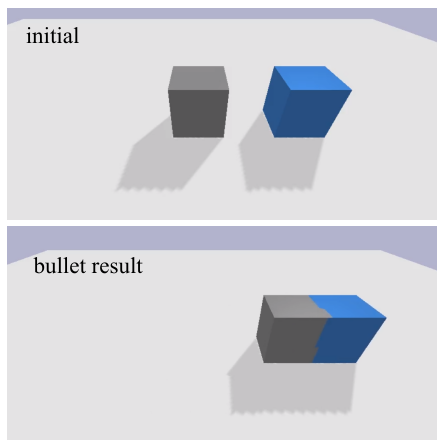}
        \label{fig:exp3_cube1}
    }
    \subfigure[Pushing cube in Nimble]
    {
        \includegraphics[height=1.8in]{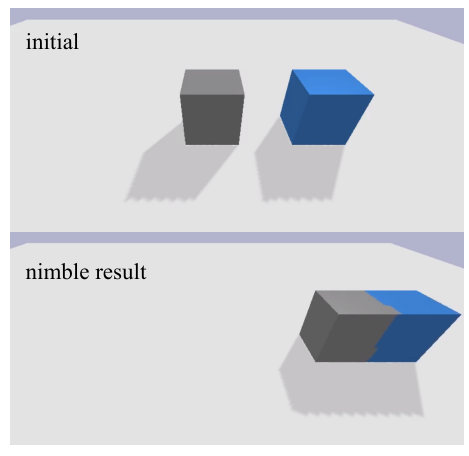}
        \label{fig:exp3_cube2}
    }
    \subfigure[Pushing cube in \textbf{Jade}]
    {
        \includegraphics[height=1.8in]{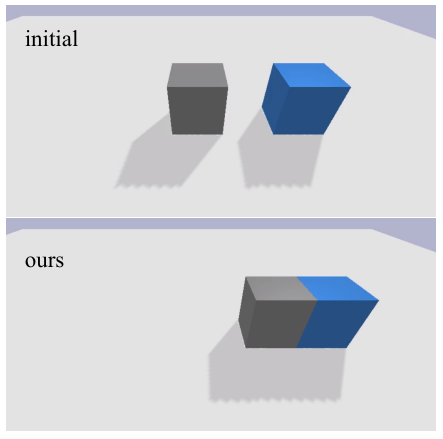}
        \label{fig:exp3_cube3}
    }
    \\
    \subfigure[The X-axis represents the time and the Y-axis represents the velocity, which represents the velocity change of the two cubes in the Bullet]
    {
        \includegraphics[width=2.5in]{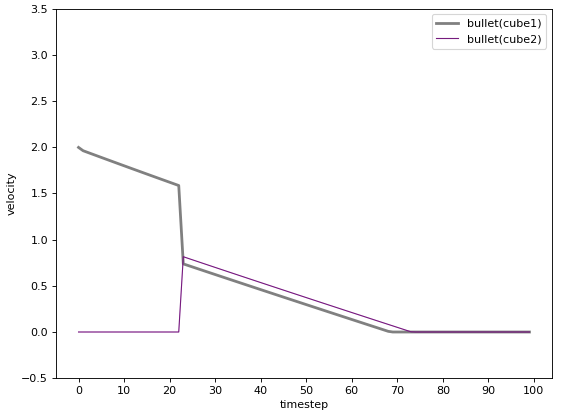}
        \label{fig:exp3_cube4}
    }
    \ \ \ \ \ \
    \subfigure[The X-axis represents the time and the Y-axis represents the velocity, which represents the velocity change of the two cubes in the Nimble and \textbf{Jade}]
    {
        \includegraphics[width=2.5in]{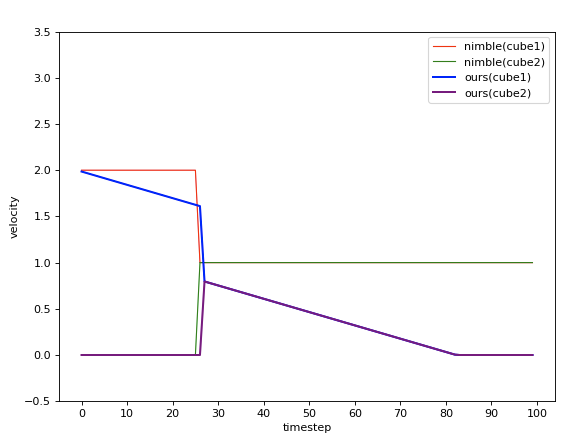}
        \label{fig:exp3_cube5}
    }
    \caption{We set the initial velocity for gray cube to hit the blue cube on the table with friction. The gray cub the blue one hits with inelastic collision and slides.}
    \label{fig:exp3_push1}
\end{figure*}

We design two types of tasks on a table with friction. The first type is that the rigid bodies of different shapes with given initial velocity slide on the table and eventually come to rest. The second type is that a cube with initial velocity slides and hits another cube and finally comes to rest together.

On the first task, we simulate different prisms sliding on a table with an initial velocity of 2$m/s$. We set each time step as 0.01 second and simulate for 150 time steps. Acceleration of gravity of the system is $9m/s^2$, and system friction coefficient is 0.16. As shown in Fig.~\ref{fig:exp3_slide1}, the slope is -0.0147, which is really close to the correct slope, -0.0144. Excluding floating point and mesh accuracy errors, our experiments correctly calculate sliding friction at multiple associated contact points. The velocity of prisms changed by time and shows in Fig.~\ref{fig:slide_table}.

We also set up a more challenging task for drop friction problem. We place one gray cube and one blue cube on the desktop at a distance of 0.5m. Set the initial velocity of the gray cube to hit the blue cube forward. Both cubes have the same mass. The rest of the physical parameters are the same as the former. The simulation result is expected and shown in Fig.~\ref{fig:exp3_push1}. 

We compare our result with Nimble and Bullet in Fig.~\ref{fig:exp3_slide1} and Fig.~\ref{fig:exp3_push1}. Since nimble's LCP Solver cannot solve the problem of multiple associated contact points, invalid LCP solution is detected and Nimble produce a solution under the friction less setting to arrive at a false solution. We solve this problem by modifying the LCP Solver and obtain the correct friction force. 

%% file: tex/con.tex
This paper introduces Jade, a differentiable physics engine for articulated rigid bodies. We use the continuous collision detection to detect the time of impact for collision checking and adopt the backtrack strategy to prevent intersection between bodies with complex geometry shapes. We derive the gradient calculation to ensure the whole simulation process differentiable under the backtrack mechanism. Our simulation models contacts as the Linear Complementarity Problem~(LCP). We modified the popular Dantzig algorithm to get valid solution under multiple frictional contacts.

%% file: root.bbl
\begin{thebibliography}{52}
\providecommand{\natexlab}[1]{#1}
\providecommand{\url}[1]{#1}
\csname url@samestyle\endcsname
\providecommand{\newblock}{\relax}
\providecommand{\bibinfo}[2]{#2}
\providecommand{\BIBentrySTDinterwordspacing}{\spaceskip=0pt\relax}
\providecommand{\BIBentryALTinterwordstretchfactor}{4}
\providecommand{\BIBentryALTinterwordspacing}{\spaceskip=\fontdimen2\font plus
\BIBentryALTinterwordstretchfactor\fontdimen3\font minus
  \fontdimen4\font\relax}
\providecommand{\BIBforeignlanguage}[2]{{%
\expandafter\ifx\csname l@#1\endcsname\relax
\typeout{** WARNING: IEEEtranN.bst: No hyphenation pattern has been}%
\typeout{** loaded for the language `#1'. Using the pattern for}%
\typeout{** the default language instead.}%
\else
\language=\csname l@#1\endcsname
\fi
#2}}
\providecommand{\BIBdecl}{\relax}
\BIBdecl

\bibitem[Paszke et~al.(2019)Paszke, Gross, Massa, Lerer, Bradbury, Chanan,
  Killeen, Lin, Gimelshein, Antiga, et~al.]{pytorch}
A.~Paszke, S.~Gross, F.~Massa, A.~Lerer, J.~Bradbury, G.~Chanan, T.~Killeen,
  Z.~Lin, N.~Gimelshein, L.~Antiga \emph{et~al.}, ``Pytorch: An imperative
  style, high-performance deep learning library,'' \emph{Advances in neural
  information processing systems}, vol.~32, pp. 8026--8037, 2019.

\bibitem[Team et~al.(2016)Team, Al-Rfou, Alain, Almahairi, Angermueller,
  Bahdanau, Ballas, Bastien, Bayer, Belikov, et~al.]{team2016theano}
T.~T.~D. Team, R.~Al-Rfou, G.~Alain, A.~Almahairi, C.~Angermueller,
  D.~Bahdanau, N.~Ballas, F.~Bastien, J.~Bayer, A.~Belikov \emph{et~al.},
  ``Theano: A python framework for fast computation of mathematical
  expressions,'' \emph{arXiv preprint arXiv:1605.02688}, 2016.

\bibitem[Hu et~al.(2019{\natexlab{a}})Hu, Li, Anderson, Ragan-Kelley, and
  Durand]{hu2019taichi}
Y.~Hu, T.-M. Li, L.~Anderson, J.~Ragan-Kelley, and F.~Durand, ``Taichi: a
  language for high-performance computation on spatially sparse data
  structures,'' \emph{ACM Transactions on Graphics (TOG)}, vol.~38, no.~6, p.
  201, 2019.

\bibitem[Bell(2020)]{bellCppAD}
B.~Bell, ``Cppad: a package for c++ algorithmic differentiation,''
  \url{http://www.coin-or.org/CppAD}, 2020.

\bibitem[Bradbury et~al.(2018)Bradbury, Frostig, Hawkins, Johnson, Leary,
  Maclaurin, Necula, Paszke, Vander{P}las, Wanderman-{M}ilne, and
  Zhang]{jax2018github}
\BIBentryALTinterwordspacing
J.~Bradbury, R.~Frostig, P.~Hawkins, M.~J. Johnson, C.~Leary, D.~Maclaurin,
  G.~Necula, A.~Paszke, J.~Vander{P}las, S.~Wanderman-{M}ilne, and Q.~Zhang,
  ``{JAX}: composable transformations of {P}ython+{N}um{P}y programs,'' 2018.
  [Online]. Available: \url{http://github.com/google/jax}
\BIBentrySTDinterwordspacing

\bibitem[Agarwal et~al.()Agarwal, Mierle, and Others]{ceres-solver}
S.~Agarwal, K.~Mierle, and Others, ``Ceres solver,''
  \url{http://ceres-solver.org}.

\bibitem[de~Avila Belbute-Peres et~al.(2018)de~Avila Belbute-Peres, Smith,
  Allen, Tenenbaum, and Kolter]{NEURIPS2018_842424a1}
\BIBentryALTinterwordspacing
F.~de~Avila Belbute-Peres, K.~Smith, K.~Allen, J.~Tenenbaum, and J.~Z. Kolter,
  ``End-to-end differentiable physics for learning and control,'' in
  \emph{Advances in Neural Information Processing Systems}, S.~Bengio,
  H.~Wallach, H.~Larochelle, K.~Grauman, N.~Cesa-Bianchi, and R.~Garnett, Eds.,
  vol.~31.\hskip 1em plus 0.5em minus 0.4em\relax Curran Associates, Inc.,
  2018. [Online]. Available:
  \url{https://proceedings.neurips.cc/paper/2018/file/842424a1d0595b76ec4fa03c46e8d755-Paper.pdf}
\BIBentrySTDinterwordspacing

\bibitem[Degrave et~al.(2019)Degrave, Hermans, Dambre,
  et~al.]{degrave2019differentiable}
J.~Degrave, M.~Hermans, J.~Dambre \emph{et~al.}, ``A differentiable physics
  engine for deep learning in robotics,'' \emph{Frontiers in neurorobotics},
  p.~6, 2019.

\bibitem[Hu et~al.(2019{\natexlab{b}})Hu, Liu, Spielberg, Tenenbaum, Freeman,
  Wu, Rus, and Matusik]{hu2019chainqueen}
Y.~Hu, J.~Liu, A.~Spielberg, J.~B. Tenenbaum, W.~T. Freeman, J.~Wu, D.~Rus, and
  W.~Matusik, ``Chainqueen: A real-time differentiable physical simulator for
  soft robotics,'' in \emph{2019 International conference on robotics and
  automation (ICRA)}.\hskip 1em plus 0.5em minus 0.4em\relax IEEE, 2019, pp.
  6265--6271.

\bibitem[Jatavallabhula et~al.(2021)Jatavallabhula, Macklin, Golemo, Voleti,
  Petrini, Weiss, Considine, Parent-L{\'e}vesque, Xie, Erleben,
  et~al.]{jatavallabhula2021gradsim}
K.~M. Jatavallabhula, M.~Macklin, F.~Golemo, V.~Voleti, L.~Petrini, M.~Weiss,
  B.~Considine, J.~Parent-L{\'e}vesque, K.~Xie, K.~Erleben \emph{et~al.},
  ``gradsim: Differentiable simulation for system identification and visuomotor
  control,'' \emph{arXiv preprint arXiv:2104.02646}, 2021.

\bibitem[Geilinger et~al.(2020)Geilinger, Hahn, Zehnder, B{\"a}cher,
  Thomaszewski, and Coros]{geilinger2020add}
M.~Geilinger, D.~Hahn, J.~Zehnder, M.~B{\"a}cher, B.~Thomaszewski, and
  S.~Coros, ``Add: analytically differentiable dynamics for multi-body systems
  with frictional contact,'' \emph{ACM Transactions on Graphics (TOG)},
  vol.~39, no.~6, pp. 1--15, 2020.

\bibitem[Du et~al.(2021)Du, Wu, Ma, Wah, Spielberg, Rus, and
  Matusik]{du2021_diffpd}
\BIBentryALTinterwordspacing
T.~Du, K.~Wu, P.~Ma, S.~Wah, A.~Spielberg, D.~Rus, and W.~Matusik, ``Diffpd:
  Differentiable projective dynamics,'' \emph{ACM Trans. Graph.}, vol.~41,
  no.~2, nov 2021. [Online]. Available: \url{https://doi.org/10.1145/3490168}
\BIBentrySTDinterwordspacing

\bibitem[Liang et~al.(2019)Liang, Lin, and Koltun]{NEURIPS2019_28f0b864}
\BIBentryALTinterwordspacing
J.~Liang, M.~Lin, and V.~Koltun, ``Differentiable cloth simulation for inverse
  problems,'' in \emph{Advances in Neural Information Processing Systems},
  H.~Wallach, H.~Larochelle, A.~Beygelzimer, F.~d\textquotesingle
  Alch\'{e}-Buc, E.~Fox, and R.~Garnett, Eds., vol.~32.\hskip 1em plus 0.5em
  minus 0.4em\relax Curran Associates, Inc., 2019. [Online]. Available:
  \url{https://proceedings.neurips.cc/paper/2019/file/28f0b864598a1291557bed248a998d4e-Paper.pdf}
\BIBentrySTDinterwordspacing

\bibitem[Qiao et~al.(2020)Qiao, Liang, Koltun, and Lin]{qiao2020scalable}
Y.-L. Qiao, J.~Liang, V.~Koltun, and M.~C. Lin, ``Scalable differentiable
  physics for learning and control,'' \emph{arXiv preprint arXiv:2007.02168},
  2020.

\bibitem[Li et~al.(2022)Li, Du, Wu, Xu, and Matusik]{li2022diffcloth}
Y.~Li, T.~Du, K.~Wu, J.~Xu, and W.~Matusik, ``Diffcloth: Differentiable cloth
  simulation with dry frictional contact,'' \emph{ACM Transactions on Graphics
  (TOG)}, vol.~42, no.~1, pp. 1--20, 2022.

\bibitem[Yu et~al.(2023)Yu, Zhao, Luo, Yang, and Shao]{yu2023diffclothai}
X.~Yu, S.~Zhao, S.~Luo, G.~Yang, and L.~Shao, ``Diffclothai: Differentiable
  cloth simulation with intersection-free frictional contact and differentiable
  two-way coupling with articulated rigid bodies,'' in \emph{2023 IEEE/RSJ
  International Conference on Intelligent Robots and Systems (IROS)}.\hskip 1em
  plus 0.5em minus 0.4em\relax IEEE, 2023.

\bibitem[Werling et~al.(2021)Werling, Omens, Lee, Exarchos, and
  Liu]{werling2021fast}
K.~Werling, D.~Omens, J.~Lee, I.~Exarchos, and C.~K. Liu, ``Fast and
  feature-complete differentiable physics for articulated rigid bodies with
  contact,'' in \emph{Proceedings of Robotics: Science and Systems (RSS)}, July
  2021.

\bibitem[Ha et~al.(2017)Ha, Coros, Alspach, Kim, and
  Yamane]{58d18a9e94704e8bb63f5e2959de61d6}
S.~Ha, S.~Coros, A.~Alspach, J.~Kim, and K.~Yamane,
  ``\BIBforeignlanguage{English (US)}{Joint optimization of robot design and
  motion parameters using the implicit function theorem},'' in
  \emph{\BIBforeignlanguage{English (US)}{Robotics}}, ser. Robotics: Science
  and Systems, S.~Srinivasa, N.~Ayanian, N.~Amato, and S.~Kuindersma,
  Eds.\hskip 1em plus 0.5em minus 0.4em\relax United States: MIT Press
  Journals, 2017, publisher Copyright: {\textcopyright} 2017 MIT Press
  Journals. All rights reserved.; 2017 Robotics: Science and Systems, RSS 2017
  ; Conference date: 12-07-2017 Through 16-07-2017.

\bibitem[Qiao et~al.(2021)Qiao, Liang, Koltun, and Lin]{qiao2021efficient}
Y.-L. Qiao, J.~Liang, V.~Koltun, and M.~C. Lin, ``Efficient differentiable
  simulation of articulated bodies,'' in \emph{International Conference on
  Machine Learning}.\hskip 1em plus 0.5em minus 0.4em\relax PMLR, 2021, pp.
  8661--8671.

\bibitem[Um et~al.(2020)Um, Brand, Fei, Holl, and Thuerey]{um2020solver}
K.~Um, R.~Brand, Y.~R. Fei, P.~Holl, and N.~Thuerey, ``Solver-in-the-loop:
  Learning from differentiable physics to interact with iterative
  pde-solvers,'' \emph{Advances in Neural Information Processing Systems},
  vol.~33, pp. 6111--6122, 2020.

\bibitem[Wandel et~al.(2020)Wandel, Weinmann, and Klein]{wandel2020learning}
N.~Wandel, M.~Weinmann, and R.~Klein, ``Learning incompressible fluid dynamics
  from scratch--towards fast, differentiable fluid models that generalize,''
  \emph{arXiv preprint arXiv:2006.08762}, 2020.

\bibitem[Holl et~al.(2020)Holl, Koltun, and Thuerey]{holl2020learning}
P.~Holl, V.~Koltun, and N.~Thuerey, ``Learning to control pdes with
  differentiable physics,'' \emph{arXiv preprint arXiv:2001.07457}, 2020.

\bibitem[Takahashi et~al.(2021)Takahashi, Liang, Qiao, and
  Lin]{Takahashi_Liang_Qiao_Lin_2021}
\BIBentryALTinterwordspacing
T.~Takahashi, J.~Liang, Y.-L. Qiao, and M.~C. Lin, ``Differentiable fluids with
  solid coupling for learning and control,'' \emph{Proceedings of the AAAI
  Conference on Artificial Intelligence}, vol.~35, no.~7, pp. 6138--6146, May
  2021. [Online]. Available:
  \url{https://ojs.aaai.org/index.php/AAAI/article/view/16764}
\BIBentrySTDinterwordspacing

\bibitem[Heiden et~al.(2021)Heiden, Millard, Coumans, Sheng, and
  Sukhatme]{heiden2021neuralsim}
\BIBentryALTinterwordspacing
E.~Heiden, D.~Millard, E.~Coumans, Y.~Sheng, and G.~S. Sukhatme, ``Neural{S}im:
  Augmenting differentiable simulators with neural networks,'' in
  \emph{Proceedings of the IEEE International Conference on Robotics and
  Automation (ICRA)}, 2021. [Online]. Available:
  \url{https://github.com/google-research/tiny-differentiable-simulator}
\BIBentrySTDinterwordspacing

\bibitem[Toussaint et~al.(2019)Toussaint, Allen, Smith, and
  Tenenbaum]{escidoc:3221511}
\BIBentryALTinterwordspacing
M.~Toussaint, K.~R. Allen, K.~A. Smith, and J.~B. Tenenbaum, ``Differentiable
  physics and stable modes for tool-use and manipulation planning \textendash
  extended abstract,'' in \emph{Proceedings of the Twenty-Eighth International
  Joint Conference on Artificial Intelligence (IJCAI-19)}.\hskip 1em plus 0.5em
  minus 0.4em\relax California: International Joint Conferences on Artificial
  Intelligence, 2019, pp. 6231--6235. [Online]. Available:
  \url{https://www.ijcai.org/Proceedings/2019/}
\BIBentrySTDinterwordspacing

\bibitem[Schenck and Fox(2018)]{schenck2018spnets}
C.~Schenck and D.~Fox, ``Spnets: Differentiable fluid dynamics for deep neural
  networks,'' in \emph{Conference on Robot Learning}.\hskip 1em plus 0.5em
  minus 0.4em\relax PMLR, 2018, pp. 317--335.

\bibitem[Holl et~al.(2019)Holl, Thuerey, and Koltun]{holl2019learning}
P.~Holl, N.~Thuerey, and V.~Koltun, ``Learning to control pdes with
  differentiable physics,'' in \emph{International Conference on Learning
  Representations}, 2019.

\bibitem[Zhu et~al.(2023)Zhu, Ke, Xu, Sun, Bai, Lv, Liu, Zeng, Ye, Lu,
  Tomizuka, and Shao]{Zhu2023DiffLfD}
X.~Zhu, J.~Ke, Z.~Xu, Z.~Sun, B.~Bai, J.~Lv, Q.~Liu, Y.~Zeng, Q.~Ye, C.~Lu,
  M.~Tomizuka, and L.~Shao, ``Diff-lfd: Contact-aware model-based learning from
  visual demonstration for robotic manipulation via differentiable
  physics-based simulation and rendering,'' in \emph{Conference on Robot
  Learning (CoRL)}, 2023.

\bibitem[Lv et~al.(2023)Lv, Feng, Zhang, Zhao, Shao, and Lu]{lv2022sam}
J.~Lv, Y.~Feng, C.~Zhang, S.~Zhao, L.~Shao, and C.~Lu, ``Sam-rl: Sensing-aware
  model-based reinforcement learning via differentiable physics-based
  simulation and rendering,'' 2023.

\bibitem[Lv et~al.(2022)Lv, Yu, Shao, Liu, Xu, and Lu]{lv2022sagci}
J.~Lv, Q.~Yu, L.~Shao, W.~Liu, W.~Xu, and C.~Lu, ``Sagci-system: Towards
  sample-efficient, generalizable, compositional, and incremental robot
  learning,'' in \emph{2022 IEEE International Conference on Robotics and
  Automation (ICRA)}.\hskip 1em plus 0.5em minus 0.4em\relax IEEE, 2022.

\bibitem[Chen et~al.(2022)Chen, Li, Lu, Fu, and Jiang]{chen2022midas}
Y.~Chen, M.~Li, W.~Lu, C.~Fu, and C.~Jiang, ``Midas: A multi-joint robotics
  simulator with intersection-free frictional contact,'' \emph{arXiv preprint
  arXiv:2210.00130}, 2022.

\bibitem[Li et~al.(2020)Li, Ferguson, Schneider, Langlois, Zorin, Panozzo,
  Jiang, and Kaufman]{Li2020IPC}
M.~Li, Z.~Ferguson, T.~Schneider, T.~Langlois, D.~Zorin, D.~Panozzo, C.~Jiang,
  and D.~M. Kaufman, ``Incremental potential contact: Intersection- and
  inversion-free large deformation dynamics,'' \emph{ACM Trans. Graph.
  (SIGGRAPH)}, vol.~39, no.~4, 2020.

\bibitem[Howell et~al.(2022)Howell, Cleac'h, Kolter, Schwager, and
  Manchester]{howell2022dojo}
T.~A. Howell, S.~L. Cleac'h, J.~Z. Kolter, M.~Schwager, and Z.~Manchester,
  ``Dojo: A differentiable simulator for robotics,'' \emph{arXiv preprint
  arXiv:2203.00806}, 2022.

\bibitem[Smith(2008)]{ode:2008}
\BIBentryALTinterwordspacing
R.~Smith, ``Open dynamics engine,'' 2008, http://www.ode.org/. [Online].
  Available: \url{http://www.ode.org/}
\BIBentrySTDinterwordspacing

\bibitem[Coumans and Bai(2016--2021)]{coumans2021}
E.~Coumans and Y.~Bai, ``Pybullet, a python module for physics simulation for
  games, robotics and machine learning,'' \url{http://pybullet.org},
  2016--2021.

\bibitem[Lee et~al.(2018)Lee, X.~Grey, Ha, Kunz, Jain, Ye, S.~Srinivasa,
  Stilman, and Karen~Liu]{lee2018dart}
J.~Lee, M.~X.~Grey, S.~Ha, T.~Kunz, S.~Jain, Y.~Ye, S.~S.~Srinivasa,
  M.~Stilman, and C.~Karen~Liu, ``Dart: Dynamic animation and robotics
  toolkit,'' \emph{The Journal of Open Source Software}, vol.~3, no.~22, p.
  500, 2018.

\bibitem[Tedrake and the Drake Development~Team(2019)]{drake}
\BIBentryALTinterwordspacing
R.~Tedrake and the Drake Development~Team, ``Drake: Model-based design and
  verification for robotics,'' 2019. [Online]. Available:
  \url{https://drake.mit.edu}
\BIBentrySTDinterwordspacing

\bibitem[phy()]{physX}


\bibitem[Todorov et~al.(2012)Todorov, Erez, and Tassa]{conf/iros/TodorovET12}
\BIBentryALTinterwordspacing
E.~Todorov, T.~Erez, and Y.~Tassa, ``Mujoco: A physics engine for model-based
  control.'' in \emph{IROS}.\hskip 1em plus 0.5em minus 0.4em\relax IEEE, 2012,
  pp. 5026--5033. [Online]. Available:
  \url{http://dblp.uni-trier.de/db/conf/iros/iros2012.html#TodorovET12}
\BIBentrySTDinterwordspacing

\bibitem[Giftthaler et~al.(2017)Giftthaler, Neunert, St{\"a}uble, Frigerio,
  Semini, and Buchli]{giftthaler2017automatic}
M.~Giftthaler, M.~Neunert, M.~St{\"a}uble, M.~Frigerio, C.~Semini, and
  J.~Buchli, ``Automatic differentiation of rigid body dynamics for optimal
  control and estimation,'' \emph{Advanced Robotics}, vol.~31, no.~22, pp.
  1225--1237, 2017.

\bibitem[Carpentier and Mansard(2018)]{Carpentier2018AnalyticalDO}
J.~Carpentier and N.~Mansard, ``Analytical derivatives of rigid body dynamics
  algorithms,'' \emph{Robotics: Science and Systems XIV}, 2018.

\bibitem[Heiden et~al.(2020)Heiden, Millard, Coumans, Sheng, and
  Sukhatme]{heiden2020neuralsim}
E.~Heiden, D.~Millard, E.~Coumans, Y.~Sheng, and G.~S. Sukhatme, ``Neuralsim:
  Augmenting differentiable simulators with neural networks,'' \emph{arXiv
  preprint arXiv:2011.04217}, 2020.

\bibitem[Freeman et~al.(2021)Freeman, Frey, Raichuk, Girgin, Mordatch, and
  Bachem]{freeman2021brax}
\BIBentryALTinterwordspacing
C.~D. Freeman, E.~Frey, A.~Raichuk, S.~Girgin, I.~Mordatch, and O.~Bachem,
  ``Brax - a differentiable physics engine for large scale rigid body
  simulation,'' in \emph{Thirty-fifth Conference on Neural Information
  Processing Systems Datasets and Benchmarks Track (Round 1)}, 2021. [Online].
  Available: \url{https://openreview.net/forum?id=VdvDlnnjzIN}
\BIBentrySTDinterwordspacing

\bibitem[Macklin(2022)]{warp2022}
M.~Macklin, ``Warp: A high-performance python framework for gpu simulation and
  graphics,'' \url{https://github.com/nvidia/warp}, March 2022, nVIDIA GPU
  Technology Conference (GTC).

\bibitem[Hecker(2004)]{lemke}
C.~Hecker, ``Lemke’s algorithm: The hammer in your math toolbox?'' Online
  slides from Game Developer Conference, 2004.

\bibitem[Cottle and Dantzig(1968)]{COTTLE1968103}
\BIBentryALTinterwordspacing
R.~W. Cottle and G.~B. Dantzig, ``Complementary pivot theory of mathematical
  programming,'' \emph{Linear Algebra and its Applications}, vol.~1, no.~1, pp.
  103--125, 1968. [Online]. Available:
  \url{https://www.sciencedirect.com/science/article/pii/0024379568900529}
\BIBentrySTDinterwordspacing

\bibitem[Baraff(1994)]{10.1145/192161.192168}
\BIBentryALTinterwordspacing
D.~Baraff, ``Fast contact force computation for nonpenetrating rigid bodies,''
  in \emph{Proceedings of the 21st Annual Conference on Computer Graphics and
  Interactive Techniques}, ser. SIGGRAPH '94.\hskip 1em plus 0.5em minus
  0.4em\relax New York, NY, USA: Association for Computing Machinery, 1994, p.
  23–34. [Online]. Available: \url{https://doi.org/10.1145/192161.192168}
\BIBentrySTDinterwordspacing

\bibitem[Coutinho(2012)]{10.5555/2392643}
M.~G. Coutinho, \emph{Guide to Dynamic Simulations of Rigid Bodies and Particle
  Systems}.\hskip 1em plus 0.5em minus 0.4em\relax Springer Publishing Company,
  Incorporated, 2012.

\bibitem[Tarbouriech and Suleiman(2018)]{bisection}
S.~Tarbouriech and W.~Suleiman, ``On bisection continuous collision checking
  method: Spherical joints and minimum distance to obstacles,'' in \emph{2018
  IEEE International Conference on Robotics and Automation (ICRA)}, 2018, pp.
  7613--7619.

\bibitem[Baraff(1989)]{collision_lcp}
\BIBentryALTinterwordspacing
D.~Baraff, ``Analytical methods for dynamic simulation of non-penetrating rigid
  bodies,'' vol.~23, no.~3, p. 223–232, jul 1989. [Online]. Available:
  \url{https://doi.org/10.1145/74334.74356}
\BIBentrySTDinterwordspacing

\bibitem[Redon et~al.(2002)Redon, Kheddar, and Coquillart]{tunneling}
S.~Redon, A.~Kheddar, and S.~Coquillart, ``Fast continuous collision detection
  between rigid bodies,'' \emph{Computer Graphics Forum}, vol.~21, 05 2002.

\bibitem[Hu et~al.(2020)Hu, Anderson, Li, Sun, Carr, Ragan-Kelley, and
  Durand]{hu2019difftaichi}
Y.~Hu, L.~Anderson, T.-M. Li, Q.~Sun, N.~Carr, J.~Ragan-Kelley, and F.~Durand,
  ``Difftaichi: Differentiable programming for physical simulation,''
  \emph{ICLR}, 2020.

\end{thebibliography}
